\documentclass[conference]{IEEEtran}
\IEEEoverridecommandlockouts
\usepackage{cite}
\usepackage{amsmath,amssymb,amsfonts}
\usepackage{algorithmic}
\usepackage{graphicx}
\usepackage{textcomp}
\usepackage{xcolor}
\def\BibTeX{{\rm B\kern-.05em{\sc i\kern-.025em b}\kern-.08em
    T\kern-.1667em\lower.7ex\hbox{E}\kern-.125emX}}

\usepackage{ntheorem}
\usepackage{enumitem}
\usepackage[caption=false,font=footnotesize]{subfig} 

\theoremseparator{.}
\theoremheaderfont{\itshape\bfseries}
\newtheorem{remark}{Remark}
\newtheorem{assumption}{Assumption}
\newtheorem{theorem}{Theorem}

\begin{document}
	
\title{Torque Control with Joints Position and Velocity Limits Avoidance}

\author{\IEEEauthorblockN{Venus Pasandi}
\IEEEauthorblockA{\textit{Artificial and Mechanical Intelligence} \\
\textit{Italian Institute of Technology}\\
Genoa, Italy \\
venus.pasandi@iit.it}
\and
\IEEEauthorblockN{Daniele Pucci}
\IEEEauthorblockA{\textit{Artificial and Mechanical Intelligence} \\
\textit{Italian Institute of Technology}\\
Genoa, Italy \\
daniele.pucci@iit.it}
}


\IEEEoverridecommandlockouts
\IEEEpubid{\makebox[\columnwidth+\columnsep+\columnwidth]{\parbox[t]{\textwidth}{\bigskip \copyright2023 IEEE. Personal use of this material is permitted. Permission from IEEE must be obtained for all other uses, in any current or future media, including reprinting/republishing this material for advertising or promotional purposes, creating new collective works, for resale or redistribution to servers or lists, or reuse of any copyrighted component of this work in other works. \hfill} \hfill}}

\maketitle

\begin{abstract}
The design of a control architecture for providing the desired motion along with the realization of the joint limitation of a robotic system is still an open challenge in control and robotics.
This paper presents a torque control architecture for fully actuated manipulators for tracking the desired time-varying trajectory while ensuring the joints position and velocity limits.
The presented architecture stems from the parametrization of the feasible joints position and velocity space by exogenous states.
The proposed parametrization transforms the control problem with constrained states to an un-constrained one by replacing the joints position and velocity with the exogenous states.
With the help of Lyapunov-based arguments, we prove that the proposed control architecture ensures the stability and convergence of the desired joint trajectory along with the joints position and velocity limits avoidance.
We validate the performance of proposed architecture through various simulations on a simple two-degree-of-freedom manipulator and the humanoid robot iCub.
\end{abstract}


\section{Introduction}
Nonlinear feedback control of unconstrained fully actuated manipulators is not new to the control community.
A large variety of position, velocity, and torque control algorithms have been developed through feedback linearization, backstepping, robust control, and adaptive tools for steering these nonlinear systems towards the desired quantities \cite{canudas1997theory}.
Although, the control algorithm is applicable/safe for a robotic manipulator when its physical constraints, such as the motion and actuation limits, are satisfied.
In this paper, we propose a control architecture to ensure the joints position and velocity limits avoidance for torque-controlled manipulators.

The joint position and velocity limits avoidance for the reference trajectory/path planning has been widely studied through neural networks \cite{zhang2003dual}, repulsive potentials \cite{dahlin2019adaptive}, optimisation \cite{almasri2021trajectory}, and parametrization~\cite{pasandi2020programmable}.
However, ensuring the physical limits for the reference trajectory does not imply that the controller's reaction to the initial conditions and disturbances will not lead to violating these limits. 

Reducing the joint velocity/acceleration of the robot considering its position and velocity limits is a simple approach for ensuring these limits.
For example, the joints acceleration can be constrained with a function of the distance between the joints position and velocity values and the corresponding limits \cite{del2017joint}.
However, these approaches rely mostly on hand-tuned and/or offline heuristics.

For handling the joint position and velocity limits of robotic systems, the control objectives can be represented as an optimization problem with inequality constraints corresponding to the joint limits.
This technique is widely used in the humanoid whole-body control and human-robot interaction control \cite{feng2014optimization}.
This technique, however, still lacks the theoretical guarantee of the stability and convergence properties associated with the evolution of the system.

The barrier Lyapunov function is recently used for handling control problems with joint/state constraints~\cite{tee2009barrier,song2018barrier}.
This function yields a value that grows to infinity when the joints approach their limits.
This method is usually complicated and needs to deal with the model inaccuracies and the conflicts between the tracking objectives and the joint limits \cite{azimi2021exponential,murtaza2021real}.

Parametrization is another technique used for handling the limitations of the system states and inputs \cite{charbonneau2016line,gazar2020jerk}.
In this technique, the feasible space of the states/inputs is parametrized by an exogenous state/input.
Thus, the constrained control problem is transformed into a non-constrained one where any control tool available in the literature can be employed.
To the best of our knowledge, the control problem via parametrization technique has not been investigated for handling both the joint position and velocity limits.

This paper proposes a torque control architecture for processing the joints position and velocity limits for fully actuated manipulators.
The proposed control provides the asymptotic stability of the desired trajectory while preserving the constant joints position and velocity limits.
Comparing to the existing methods for implementing joints position and velocity limits in torque-controlled manipulators, our proposed architecture, with the help of Lyapunov arguments, ensures that the desired trajectory is asymptotically stable and the time evolution of the joints position and velocity always remains within the associated limits provided that the desired trajectory satisfies these limits. 
For this purpose, the feasible joint position and velocity space is parametrized by exogenous states.
The proposed parametrization introduces a one-to-one map between the joints position and velocity of the robot and the exogenous states.
Using the exogenous states, the problem of control design with constrained states (i.e. constrained joints position and velocity) is transformed to an un-constrained control problem with exogenous states.
After that, a control policy is proposes for ensuring the stability properties of the exogenous states which leads us to ensuring the stability properties and joints limitation avoidance of the robot.
We have investigated the performance and limitations of the proposed control architecture through simulations on a simple two-degree-of-freedom manipulator, and also humanoid robot iCub \cite{metta2008icub}.

This paper is organized as follows.
Section~\ref{sec:notations} expresses the notation and the definitions used in the paper.
Section~\ref{sec:problemStatement} introduces the problem statements.
In section~\ref{sec:controlArchitecture}, a control architecture is proposed for stabilizing a desired joint trajectory and ensuring the joints position and velocity limits avoidance.
In section~\ref{sec:simulationResults}, the simulation results carried out to validate the performance of the proposed control architecture are illustrated.
Finally, section~\ref{sec:conclusion} concludes the paper by remarks and perspectives.
\section{NOTATION}\label{sec:notations}

The following notation is used throughout the paper.
\begin{itemize}
	\item $\mathbb{R}$ is the set of real numbers.
	\item $I_n$ is the $n \times n$ identity matrix.
	\item For a vector $\boldsymbol{q} \in \mathbb{R}^n $, the $i^{th}$ component of $\boldsymbol{q}$ is written as $q_i$.
	\item The transpose operator is denoted by $(\cdot)^{\top}$.
	\item For a vector $\boldsymbol{q} \in \mathbb{R}^n$, the diagonal matrix of $\boldsymbol{q}$ is written as $\breve{\boldsymbol{q}}$.
	\item For a vector $\boldsymbol{q} \in \mathbb{R}^n$, the Euclidean norm of $\boldsymbol{q}$ is denoted by $\| \boldsymbol{q} \|$.
	\item For a vector $\boldsymbol{q} \in \mathbb{R}^n$, the absolute value vector of $\boldsymbol{q} $ is denoted by $| \boldsymbol{q} | $.
	\item For a vector $\boldsymbol{q} \in \mathbb{R}^n$, the function $\tanh (\boldsymbol{q}) : \mathbb{R}^n \rightarrow \mathbb{R}^n$ is defined as $\tanh(\boldsymbol{q}) = \left[ \tanh (q_1), \tanh (q_2), ..., \tanh (q_n) \right]^{\top} $.
	\item For a scalar $q \in \mathbb{R}$, the function $\text{Sat} (q,q_{min},q_{max}): \mathbb{R} \rightarrow \mathbb{R}$ is defined as
	\begin{equation*}
		\textmd{Sat} (q,q_{min},q_{max}) =
		\begin{cases}
		q_{max} & q \geq q_{max} \\
		q & q_{min} \leq q \leq q_{max} \\
		q_{min} & q \leq q_{min}
		\end{cases}
	\end{equation*}
	\item For a vector $\boldsymbol{q} \in \mathbb{R}^n$, the function $\text{Sat} (\boldsymbol{q},\boldsymbol{q}_{min},\boldsymbol{q}_{max}) : \mathbb{R}^n \rightarrow \mathbb{R}^n$ is defined as $\text{Sat}(\boldsymbol{q},\boldsymbol{q}_{min},\boldsymbol{q}_{max}) = \left[ \text{Sat} (q_1,q_{min_1},q_{max_1}), ..., \text{Sat} (q_n,q_{min_n},q_{max_n}) \right]^{\top} $.
\end{itemize}
\section{PROBLEM STATEMENT}\label{sec:problemStatement}

The equations of motion for a fully actuated robotic manipulator with $n$ degrees of freedom can be written in the following form \cite{sicilianoando2008khatib}
\begin{equation}\label{eq:originalEOM}
M(\boldsymbol{q}) \ddot{ \boldsymbol{q} } + C(\boldsymbol{q},\dot{\boldsymbol{q}})\dot{\boldsymbol{q}} + G(\boldsymbol{q}) = \boldsymbol{\tau},
\end{equation}
where $ \boldsymbol{q} \in \mathbb{R}^{n \times 1} $ is the vector of generalized coordinates, $ M(q) \in \mathbb{R}^{n \times n}$ is the inertia matrix, $C(\boldsymbol{q},\dot{ \boldsymbol {q} }) \dot{ \boldsymbol {q} } \in \mathbb{R}^{n}$ is the vector of the Centrifugal and Coriolis effects, and $G(q) \in \mathbb{R}^{n}$ is the vector of the Gravitational effects.
$ \tau \in \mathbb{R}^{n}$ is the vector of the actuator forces/torques.

The control objective is defined as computing $ \boldsymbol{\tau} $ such that $\boldsymbol{q}$ tracks the desired trajectory (e.g. $ \boldsymbol{q}_d(t) $).
Though, $\tau$ is feasible if it preserves the physical limits of the robot such as joint position, velocity, and torque limits.
In the present paper, we propose a control policy that ensures 
\begin{enumerate}
	\item the asymptotic stability of the desired trajectory, and
	\item the joint position and velocity limit avoidance.
\end{enumerate}

Assume that the feasible region for the generalised coordinates is as
\begin{equation}
\mathcal{Q}_q := \left \{ \boldsymbol{q} \in \mathbb{R}^{n} : { \boldsymbol {q} } _ {min} < \boldsymbol {q} < { \boldsymbol {q} } _ {max} \right \} ,
\end{equation}
where $ {\boldsymbol{q}}_{min},{\boldsymbol{q}}_{max} \in \mathbb{R}^{n} $ denote the vectors that define the minimum and maximum values for generalised coordinates.
Moreover, assume that the feasible region for the generalised velocities is as
\begin{equation}
\mathcal{Q}_{ \dot { q } } := \left\{ \dot{ \boldsymbol {q} } \in \mathbb{R}^{n} : \dot {\boldsymbol{q} }_{min} < \dot {\boldsymbol{q} } < \dot {\boldsymbol {q}}_{max} \right\},
\end{equation}
where $ \dot {\boldsymbol{q} }_{min}, \dot {\boldsymbol{q} }_{max} \in \mathbb{R}^{n} $ denotes the vector of the minimum and maximum values for generalised velocities. 

We propose a control architecture ensuring that $\boldsymbol{q}$ tracks the desired trajectory $\boldsymbol{q}_d(t)$ while the evolution of $ \boldsymbol{q} $ and $\dot{\boldsymbol{q}}$ always remain within $\mathcal{Q}_q$ and $\mathcal{Q}_{\dot{q}}$, respectively.

Throughout this paper, we assume that:
\begin{assumption}
	The first and second-order time derivatives of $\boldsymbol{q}_d$ are well-defined and bounded $\forall t \in \mathbb{R}^+$.
	Moreover, the desired trajectory $\boldsymbol{q}_d$ is feasible, i.e. $\left(\boldsymbol{q}_d(t),\dot{\boldsymbol{q}}_d(t)\right) \in \left(\mathcal{Q}_q,\mathcal{Q}_{\dot{q}}\right), \, \forall t \geq 0$. 
\end{assumption}

\begin{assumption}
	Each generalized coordinate possesses a free motion domain different from zero i.e. $q_{{max}_i} - q_{{min}_i} > 0$ and $\dot{q}_{{max}_i}-\dot{q}_{{min}_i} > 0, \, \forall i=1:n$.
\end{assumption}

\section{CONTROLLER ARCHITECTURE}\label{sec:controlArchitecture}

We parametrize the space of the generalised coordinates, as well as, the generalised velocities as follows
\begin{equation}\label{eq:forward_state_transformation}
\begin{aligned} 
\boldsymbol{q} &= \boldsymbol{q}_0 + \breve{\boldsymbol{\delta}}_{q} \tanh ( \boldsymbol{\zeta} ), \\
\dot{\boldsymbol{q}} &= \dot{\boldsymbol{q}}_0 + \breve{\boldsymbol{\delta}}_{\dot{q}} \tanh ( \boldsymbol{\psi} ), 
\end{aligned}
\end{equation}
where
\begin{equation*}
\begin{aligned}
\boldsymbol{q}_0 &= \frac{ \boldsymbol{q}_{max} + \boldsymbol{q}_{min} }{2}, \\
\boldsymbol{q}_0 &= \frac{ \dot{\boldsymbol{q}}_{max} + \dot{\boldsymbol{q}}_{min} }{2}, \\
\boldsymbol{\delta}_q &= \frac{\boldsymbol{q}_{max}-\boldsymbol{q}_{min}}{2}, \\
\boldsymbol{\delta}_{\dot{q}} &= \frac{\dot{\boldsymbol{q}}_{max} - \dot{\boldsymbol{q}}_{min}}{2}. 
\end{aligned}
\end{equation*}
The above parametrization guarantees that $ (\boldsymbol{q},\dot{\boldsymbol{q}}) \in \left( Q_q , Q_{\dot{q}} \right) $ for bounded $(\boldsymbol{\zeta} , \boldsymbol{\psi})$.
This parametrization is a one-to-one nonlinear map between $ (\boldsymbol{q},\dot{\boldsymbol{q}}) $ and $ (\boldsymbol{\zeta},\boldsymbol{\psi}) $.
The new states $(\boldsymbol{\zeta} , \boldsymbol{\psi})$ is computed according to $(\boldsymbol{q},\dot{\boldsymbol{q}})$ as
\begin{equation}\label{eq:backward_state_transformation}
\begin{aligned}
\boldsymbol{\zeta} &= \tanh^{-1} \left( \breve{\boldsymbol{\delta}}_{q}^{-1} \left( \boldsymbol{q} - \boldsymbol{q}_0 \right) \right), \\
\boldsymbol{\psi} &= \tanh^{-1} \left( \breve{\boldsymbol{\delta}}_{\dot{q}}^{-1} \left( \dot{\boldsymbol{q}} - \dot{\boldsymbol{q}}_0 \right) \right). 
\end{aligned}
\end{equation}

The second time derivative of the generalized coordinates can be computed with respect to $(\boldsymbol{ \zeta},\boldsymbol{\psi})$ as
\begin{equation}
\ddot{\boldsymbol{q}} = J_{\psi} \dot{\boldsymbol{\psi}}, 
\end{equation}
where $ J_{\psi} = \breve{\boldsymbol{\delta}}_{\dot{q}} \left( I_n - \breve{\tanh}^2 ( \boldsymbol{\psi} ) \right) $.
Thus, \eqref{eq:originalEOM} can be written in terms of $(\boldsymbol{\zeta},\boldsymbol{\psi})$ as
\begin{equation}\label{eq:transformed_dynamics}
\begin{cases}
J_{\zeta} \dot{ \boldsymbol{\zeta} } = \breve{\boldsymbol{\delta}}_{\dot{q}} \tanh(\boldsymbol{\psi}) + \dot{\boldsymbol{q}}_0 \\
M( \boldsymbol{ \zeta} ) J_{\psi} \dot{ \boldsymbol{\psi} } + h_{\zeta \psi} ( \boldsymbol{\zeta} , \boldsymbol{\psi} ) = \boldsymbol{\tau},
\end{cases}
\end{equation} 
where $ J_{\zeta} = \breve{\boldsymbol{\delta}}_{q} \left( I_n - \breve{\tanh}^2(\boldsymbol{\zeta}) \right) $ and $h_{\zeta \psi} = C(\boldsymbol{\zeta},\boldsymbol{\psi}) \left( \breve{\boldsymbol{\delta}}_{\dot{\boldsymbol{q}}} \tanh(\boldsymbol{\psi}) + \dot{\boldsymbol{q}}_0 \right) + G(\boldsymbol{\zeta})$.

\begin{remark}
One can observe the following properties:
\begin{enumerate}[label=R1.\arabic*, leftmargin=25pt]
	\item The matrices $J_\zeta$, $J_\psi$, $\dot{J}_\zeta$, and $\dot{J}_\psi$ are diagonal,
	\item For all $\boldsymbol{\zeta}$, $0 \leq J_\zeta \leq \boldsymbol{\delta}_{q}$,
	\item For all $\boldsymbol{\psi}$, $0 \leq J_\psi \leq \boldsymbol{\delta}_{\dot{q}}$,
	\item The matrix $J_\zeta$ is positive definite for bounded $\boldsymbol{\zeta}$ (i.e. for $\boldsymbol{q} \in \mathcal{Q}_q$),
	\item The matrix $J_\psi$ is positive definite for bounded $\boldsymbol{\psi}$ (i.e. for $\dot{ \boldsymbol {q} } \in \mathcal{Q}_{\dot{q}}$),
	\item The matrix $ M J_\psi $ is positive definite for bounded $\boldsymbol{\psi}$.
\end{enumerate}
\end{remark}

If the desired trajectory is feasible i.e. $ \left( \boldsymbol{q}_d , \dot{\boldsymbol{q}}_d \right) \in \left( \mathcal{Q}_q , \mathcal{Q}_{\dot{q}} \right) $, the desired trajectory of $(\boldsymbol{ \zeta},\boldsymbol{\psi})$ is defined as 
\begin{equation}\label{eq:desired_new_trajectories}
\begin{aligned}
\boldsymbol{\zeta}_d &= \tanh^{-1} \left(\boldsymbol{\delta}_{q}^{-1} \left( \boldsymbol{q}_d - \boldsymbol{q}_0 \right) \right), \\
\boldsymbol{\psi}_d &= \tanh^{-1} \left( \boldsymbol{\delta}_{\dot{q}}^{-1} \left( \dot{\boldsymbol{q}}_d - \dot{\boldsymbol{q}}_0 \right) \right). 
\end{aligned}
\end{equation}

For ensuring asymptotic stability of $ (\boldsymbol{\zeta}_d,\boldsymbol{\psi}_d) $, one can use any control technique from the literature.
The control policy is required to ensure that $(\boldsymbol{\zeta},\boldsymbol{\psi})$ is bounded and converge to $(\boldsymbol{\zeta}_d,\boldsymbol{\psi}_d)$. 
We use the feedback linearisation technique and consider the control policy as
\begin{equation}\label{eq:control_policy}
\boldsymbol{\tau} = h_{\zeta \psi} + M J_{\psi} \left( {\dot{\psi}}_r - k_2^{-1} k_1 \dot{e}_{\zeta} - k_2 e_{\zeta} - k_3 e_{\psi} \right),
\end{equation} 
where
\begin{equation*}
\begin{aligned}
e_{\zeta} &= \boldsymbol{\zeta} - \boldsymbol{\zeta}_d, \\
e_{\psi} & = \boldsymbol{\psi} - \boldsymbol{\psi}_r, \\
\boldsymbol{\psi}_r &= \tanh^{-1} \left( \breve{\boldsymbol{\delta}}_{\dot{q}}^{-1} \left( J_{\zeta} J_{\zeta_d}^{-1} \left( \breve{\boldsymbol{\delta}}_{\dot{q}} \tanh ( \boldsymbol{\psi}_d ) + \dot{\boldsymbol{q}}_0 \right) - \dot{\boldsymbol{q}}_0 \right) \right).
\end{aligned}
\end{equation*}
and $k_1,k_2,k_3 \in \mathbb{R}^n$ are constant matrices.

\begin{theorem}\label{thm:stability}
	Considering the equations of motion presented in \eqref{eq:transformed_dynamics} and the control policy \eqref{eq:control_policy}, given that
	\begin{enumerate}[label=T1.\arabic*), leftmargin=30pt]
		\item $k_1,k_2,k_3$ are diagonal and positive definite, 
		\item $k_3 - k_2 k_1^{-1}k_2$ is positive definite, and
		\item Assumption 1 and 2 hold,
		\item $\boldsymbol{\psi}_r$ is well-defined,
	\end{enumerate}
then $(\boldsymbol{\zeta}_d,\boldsymbol{\psi}_d) $ is the globally asymptotically stable trajectory of the closed loop system.
\end{theorem}

In the lights of the above theorem, if T1.1 to T1.4 are satisfied, then $(\boldsymbol{\zeta }, \boldsymbol{\psi})$ is bounded and converges to $\left( \boldsymbol{\zeta }_d,\boldsymbol{\psi}_d \right)$.
Considering \eqref{eq:forward_state_transformation}, the boundedness of $(\boldsymbol{\zeta }, \boldsymbol{\psi})$ ensures that $\left(\boldsymbol{q},\dot{ \boldsymbol {q} }\right) \in \left(\mathcal{Q}_q,\mathcal{Q}_{\dot{ q }}\right)$.
On the other hand, considering \eqref{eq:backward_state_transformation} and \eqref{eq:desired_new_trajectories}, the fact that $(\boldsymbol{\zeta }, \boldsymbol{\psi})$ converges to $\left( \boldsymbol{\zeta }_d,\boldsymbol{\psi}_d \right)$ implies that $\left(\boldsymbol{q},\dot{ \boldsymbol {q} }\right)$ converges to $\left(\boldsymbol{q}_d,\dot{\boldsymbol{q}}_d\right)$.
Note that T1.4 is satisfied in a neighborhood of the desired trajectory where the error between $\boldsymbol{ \zeta}$ and $\boldsymbol{ \zeta}_d$ is sufficiently small.
Thus, the proposed control architecture ensures the local asymptotic stability of the desired trajectory if T1.1 to T1.3 are satisfied.
Moreover, T1.4 is ensured for the initial conditions satisfying it.
However, this assumption could be violated initially due to the initial conditions or during the motion because of disturbances.
Remark 5 proposes a trick for dealing with these situations.
\section{SIMULATION RESULTS}\label{sec:simulationResults}

\begin{figure}[!t]
	\centering
	\subfloat[Two-link manipulator]{\includegraphics[width=0.42\linewidth,height=0.42\linewidth]{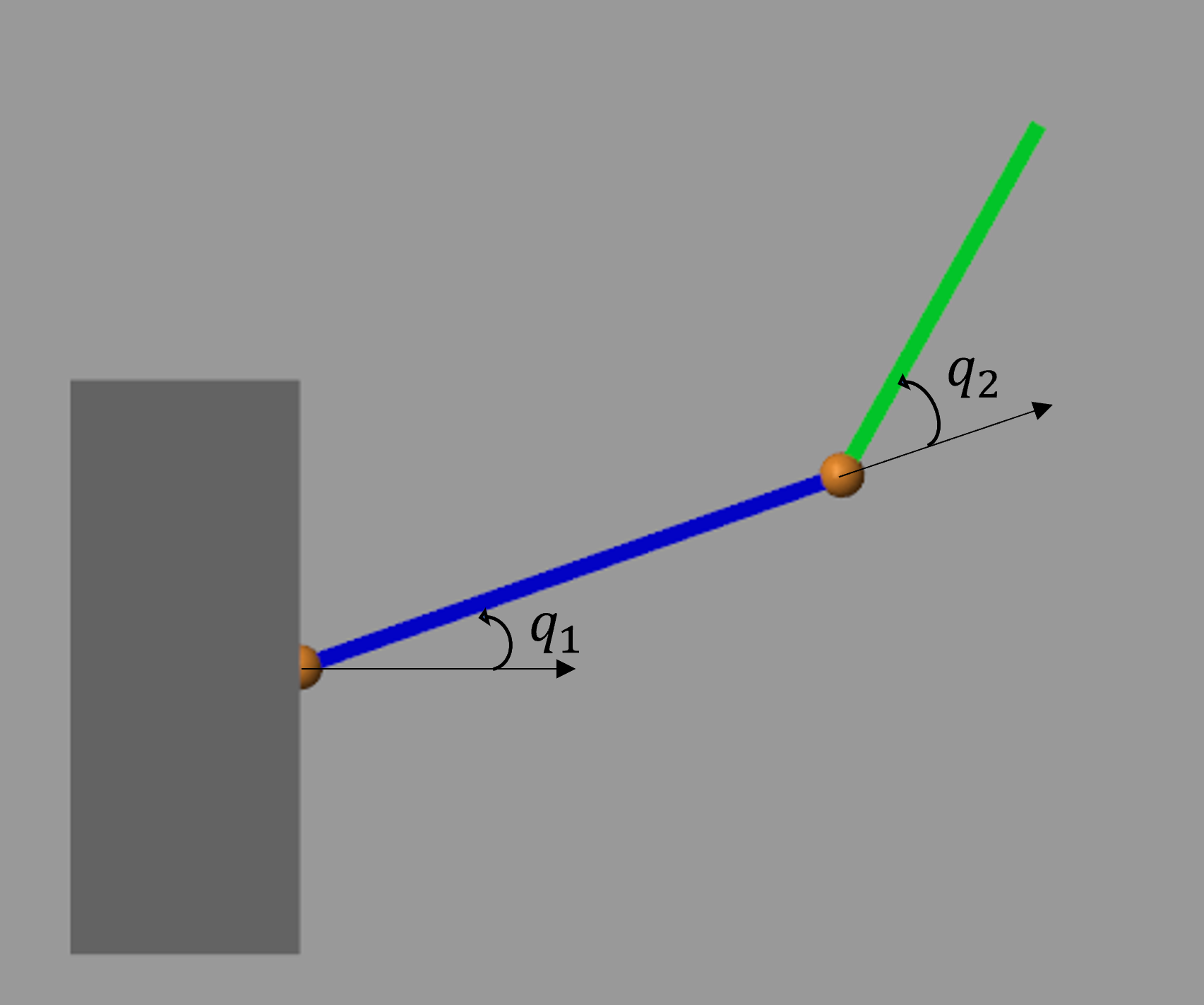}
	\label{fig:simulation:two-link-manipulator}
	}
	\hspace{1mm}
	\subfloat[iCub]{\includegraphics[width=0.40\linewidth,height=0.42\linewidth]{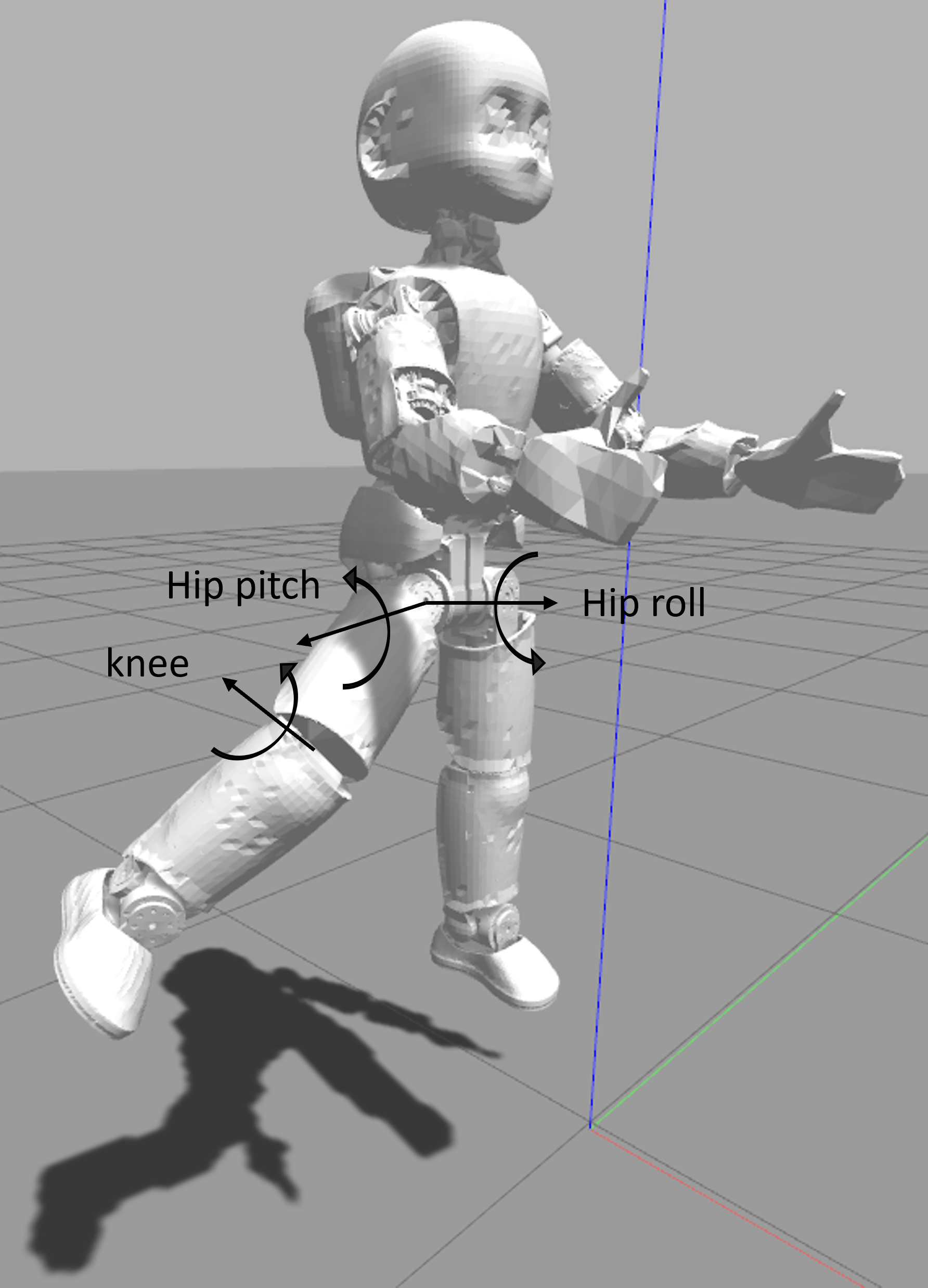}
	\label{fig:simulation:icub}
	}
	\caption{The robotic manipulators used in the simulation studies}
	\label{fig:simulation:robots}
\end{figure}

In this section, we evaluates the performance of the proposed control architecture for trajectory tracking in two simulation studies: a simple two-link manipulator and humanoid robot iCub \cite{metta2008icub} (see Fig.~\ref{fig:simulation:robots}).
In the first case, the proposed controller is used for controlling a two-degree-of-freedom manipulator simulated in Simscape/MATLAB \cite{miller2021simscape}.
In the second case, the proposed controller is employed on the humanoid robot iCub simulated in Gazebo \cite{koenig2004design}.
In both cases, the performance of the proposed control architecture is compared to the joints limit avoidance torque control (JLATC)~\cite{charbonneau2016line}.
All results are available on GitHub\footnote{https://github.com/ami-iit/paper\_pasandi\_2023\_icra-joint-limit-avoidance}.

For implementing the proposed control architecture, we need some considerations:
\begin{remark}
	Theorem \ref{thm:stability} assumes that the controller is continuous.
	In both simulation and real experiments, however, the implementation of the controller is a discrete system.
	Thus, the controller is required to be implemented by an appropriate time step for achieving the results explained in Theorem \ref{thm:stability}.
	The appropriate time step depends on the system dynamics and the desired trajectory.
	But, a small time step drastically slows down the simulations, and is not always achievable in real applications.
	To resolve the requirement of small time steps, we modify the control policy as the following
	{\small\begin{equation}
		\boldsymbol{\tau} = h_{\zeta \psi} + M J_{\psi} \left( {\dot{\psi}}_r - k_2^{-1} k_1 \dot{e}_{\zeta} - k_2 e_{\zeta} - k_3 J_{\psi}^{-1} e_{\psi} \right).
		\end{equation}}
\end{remark}

\begin{remark}
	In practice, the noise signals can cause the joints position and velocity of the system to go beyond their limits.
	In this case, the definition of $(\zeta,\psi)$ is not valid.
	To resolve this issue, we modify \eqref{eq:backward_state_transformation} as
	\begin{equation}
	\begin{aligned}
	\boldsymbol{\zeta} &= \tanh^{-1} \left( \breve{\boldsymbol{\delta}}_{q}^{-1} \left( \text{Sat} \left( \boldsymbol{q},\boldsymbol{q}_{min},\boldsymbol{q}_{max} \right) - \boldsymbol{q}_0 \right) \right), \\
	\boldsymbol{\psi} &= \tanh^{-1} \left( \breve{\boldsymbol{\delta}}{\dot{q}}^{-1} \left( \text{Sat} (\dot{\boldsymbol{q}},\dot{\boldsymbol{q}}_{min},\dot{\boldsymbol{q}}_{max}) - \dot{\boldsymbol{q}}_0 \right) \right).
	\end{aligned}
	\end{equation}
\end{remark}

\begin{remark}
	In practice, it can happen that the desired joints trajectory does not satisfy the joints position and velocity limits of the system.
	In this case, the definition of $(\zeta_d,\psi_d)$ are not valid.
	To resolve this issue, we modify the definition of $(\zeta,\psi)$ as the following
	\begin{equation}
	\begin{aligned}
	\boldsymbol{\zeta}_d &= \tanh^{-1} \left( \breve{\boldsymbol{\delta}}_{q}^{-1} \left( \text{Sat} \left( \boldsymbol{q}_d,\boldsymbol{q}_{min},\boldsymbol{q}_{max} \right) - \boldsymbol{q}_0 \right) \right), \\
	\boldsymbol{\psi}_d &= \tanh^{-1} \left( \breve{\boldsymbol{\delta}}_{\dot{q}}^{-1} \left( \text{Sat} (\dot{\boldsymbol{q}}_d,\dot{\boldsymbol{q}}_{min},\dot{\boldsymbol{q}}_{max}) - \dot{\boldsymbol{q}}_0 \right) \right).
	\end{aligned}
	\end{equation}
\end{remark}

\begin{remark}
	The definition of $\psi_r$ is well-defined if $\left| \breve{\boldsymbol{\delta}}_{\dot{q}}^{-1} \left( J_{\zeta} J_{\zeta_d}^{-1} \left( \breve{\boldsymbol{\delta}}_{\dot{q}} \tanh ( \boldsymbol{\psi}_d ) + \dot{\boldsymbol{q}}_0 \right) - \dot{\boldsymbol{q}}_0 \right) \right| \leq 1$.
	This condition can be violated because of the error between $\zeta_d$ and $\zeta$.
	Thus, we modify the definition of $\psi_r$ as
	\begin{equation}
	\psi = \tanh^{-1} \left( \dfrac{\gamma}{\left( I_n + \breve{| \gamma |}^p \right)^{1/p}} \right),
	\end{equation}
	where $\gamma = \breve{\boldsymbol{\delta}}_{\dot{q}}^{-1} \left( J_{\zeta} J_{\zeta_d}^{-1} \left( \breve{\boldsymbol{\delta}}_{\dot{q}} \tanh ( \boldsymbol{\psi}_d ) + \dot{\boldsymbol{q}}_0 \right) - \dot{\boldsymbol{q}}_0 \right)$ and $p \in \mathbb{R}$ is a positive constant. 
\end{remark}

\subsection{Two-link manipulator}
\label{sec:simulation:two-link-manipulator}

In this study, the two-link manipulator shown in Fig.\ref{fig:simulation:two-link-manipulator} is simulated in Simscape/MATLAB \cite{miller2021simscape}.
The two-link manipulator has two degrees of freedom and moves in the vertical plane.
The simulation is performed in Simulink/MATLAB with the continuous integrator "ode45" and the time step of $0.01 [sec]$.

The feasible joint position range is equal to $(-45,90) [deg]$ and $(-90,90) [deg]$, and the feasible joint velocity range is equal to $(-90,90) [deg/sec]$ and $(-90,180) [deg/sec]$ for $q_1$ and $q_2$, respectively .

\subsubsection{Constant desired trajectory}

\begin{figure}[!t]
	\centering
	\includegraphics[scale=0.18]{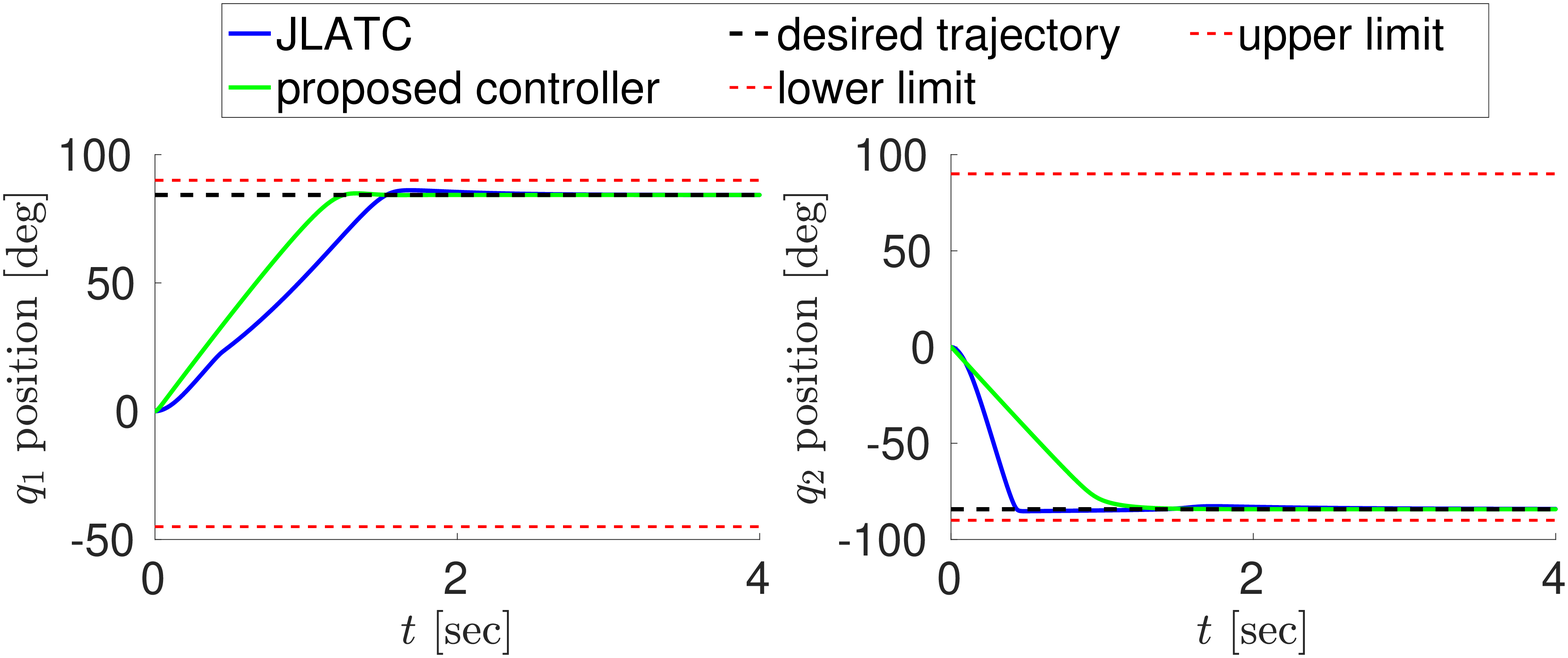}
	\includegraphics[scale=0.18]{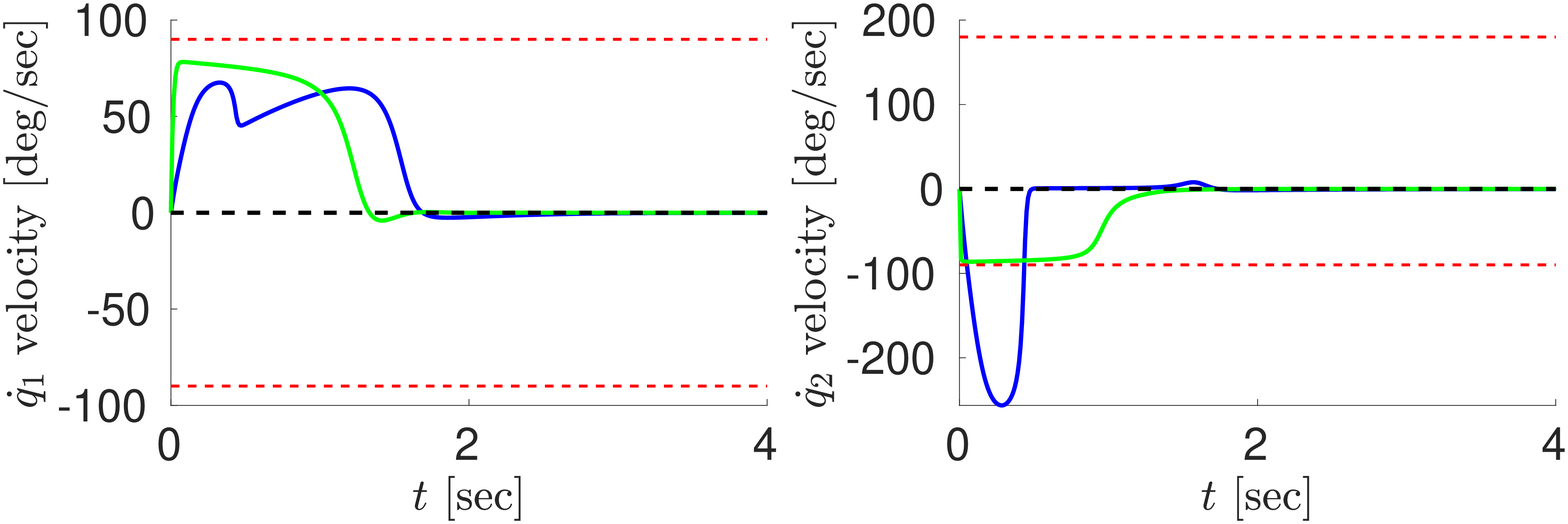}
	\includegraphics[scale=0.18]{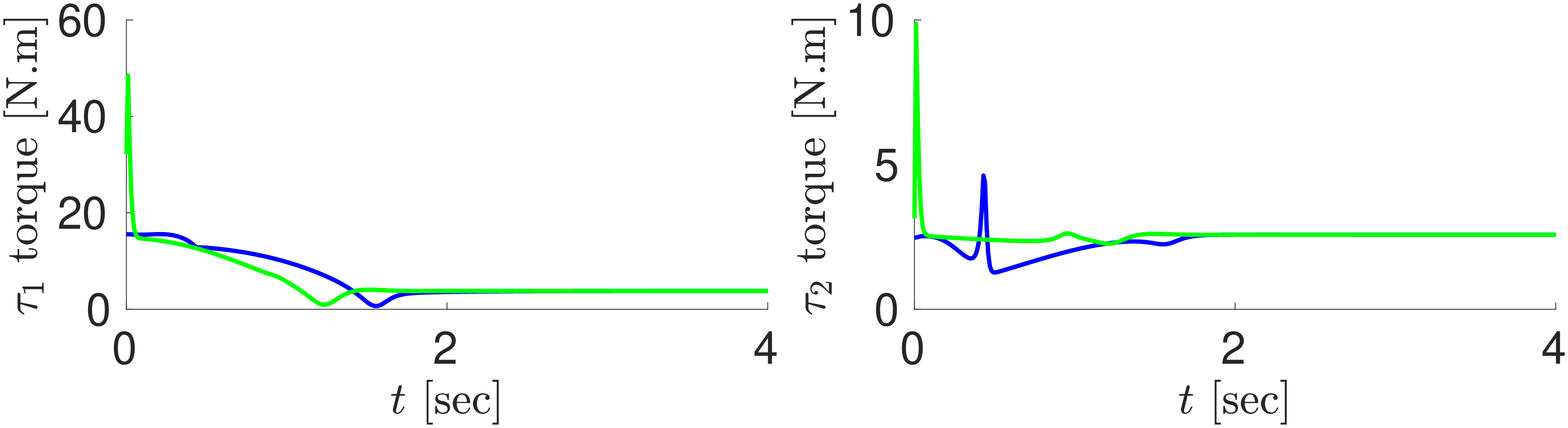}
	\caption{The time evolution of the joints position, velocity and torque of the two link manipulator in the constant desired trajectory scenario.
	The coefficients for the proposed controller are $k_1 = \text{diag}(22,505), k_2 = \text{diag} (20,50)$, and $k_3 = \text{diag}(10,5)$.
	The coefficients for the JLATC are $k_p = \text{diag}(0.4,0.1)$ and $\text{diag}(0.2,0.1)$.}
	\label{fig:simulation:manip:constant}
\end{figure}

In this scenario, the robot reaches the constant desired joint position $(90-2.5,-90+2.5) [deg]$ from the initial joint position $(0,0) [deg]$.
The results are shown in Fig.~\ref{fig:simulation:manip:constant}.
As can be seen, using the JLATC, the joint trajectory respects the predefined position limits but goes beyond the velocity limits.
Instead, using the proposed control architecture, the joints trajectory converges to the desired value while respecting the predefined position and velocity limits irrespective of the large control coefficients.
The control coefficients for the proposed control architecture are tuned for each joint to provide a convergence rate equal or higher than the one provided by the JLATC.
In this simulation, the convergence rate provided for the second joint by the proposed control architecture is the maximum allowable convergence rate considering the corresponding joint velocity limit. 

\begin{figure}[!b]
	\centering
	\includegraphics[scale=0.18]{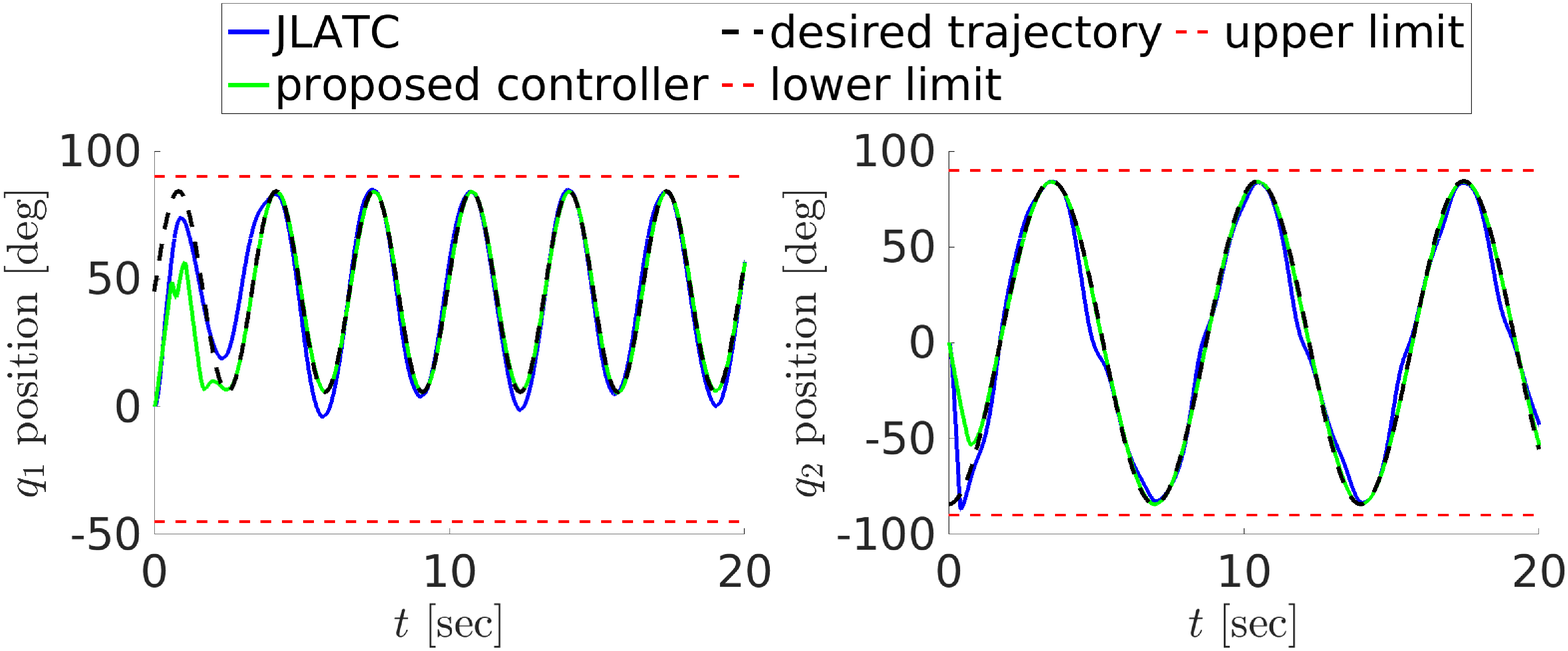}
	\includegraphics[scale=0.18]{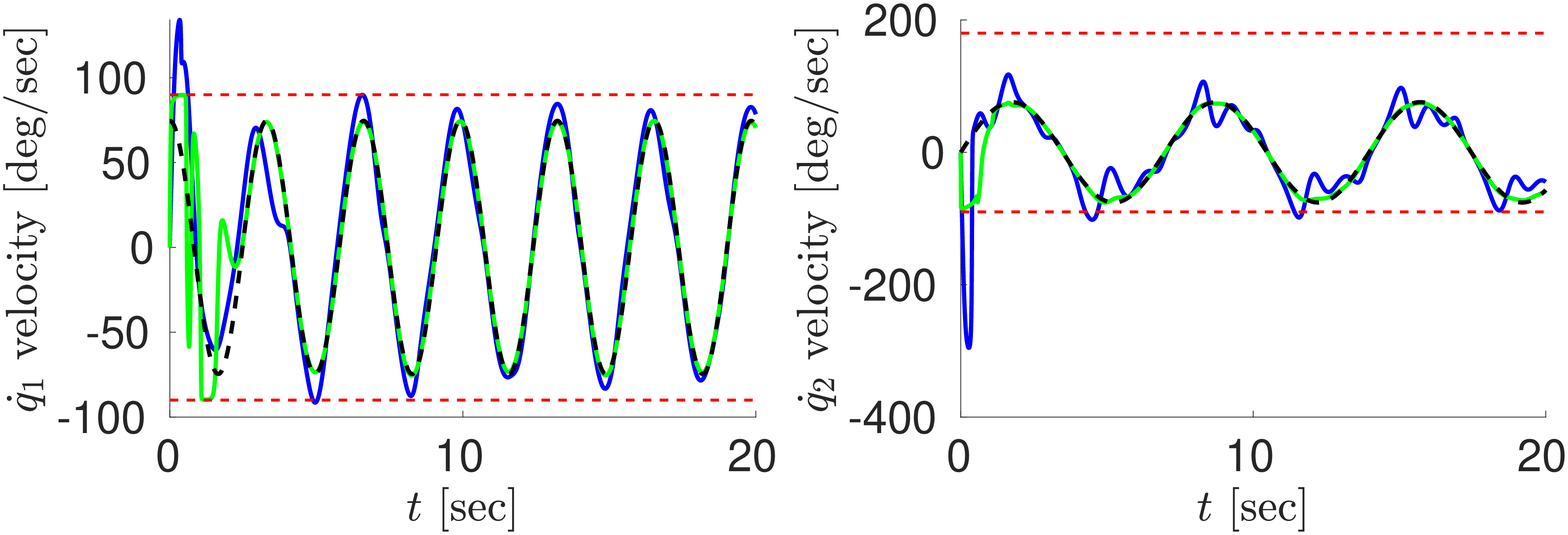}
	\includegraphics[scale=0.18]{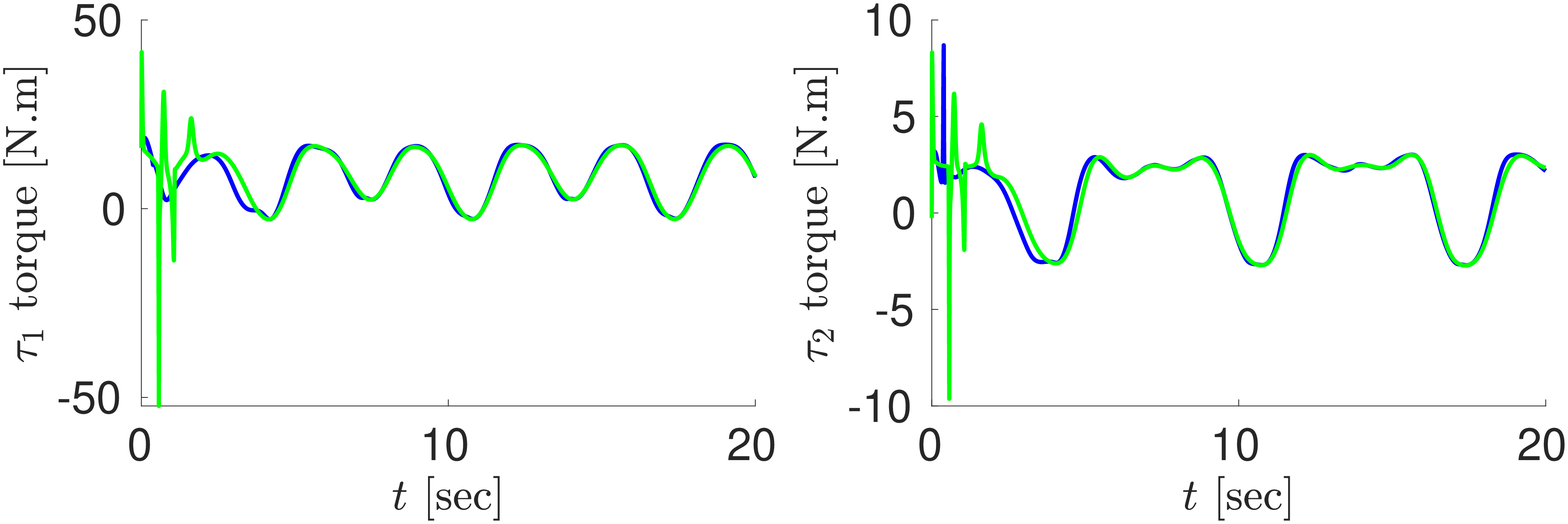}
	\caption{The joints position, velocity and torque evolution of the two link manipulator in the tracking of the sinusoidal desired trajectory.
		The coefficients for the proposed controller are $k_1 = \text{diag}(22,505), k_2 = \text{diag} (20,50)$, and $k_3 = \text{diag}(10,5)$.
		The coefficients for the JLATC are $k_p = \text{diag}(0.8,0.3)$ and $\text{diag}(0.2,0.06)$.}
	\label{fig:simulation:manip:sinusoidal}
\end{figure}

\subsubsection{Sinusoidal desired trajectory}

In this scenario, the robot tracks the sinusoidal desired trajectory $\boldsymbol{q} = A \sin \left(\omega t + \varphi_0 \right) + b$ where $A = (45-5,-90+5) [deg]$, $\omega = (1.9,0.9) [rad/sec]$, $\varphi_0 = (0,\pi/2) [rad]$, and $b = (45,0) [deg]$.
The initial position of the robot is $(0,0) [deg]$.
As shown in Fig.~\ref{fig:simulation:manip:sinusoidal}, using the proposed control architecture, the joints trajectory tracks the desired trajectory while respecting the predefined joints position and velocity limits.
Instead, using the JLATC controller, the joints trajectory respects the joints position limits but not the joint velocity limits.

\subsection{Humanoid robot iCub}

In this study, the control architecture and iCub are simulated in MATLAB and Gazebo, respectively, and communicate together through Yarp channels.
The MATLAB discrete integrator with the time step of $0.001 [sec]$ is used as the simulation integrator.

The leg of the iCub is used for this study, forming a 3 degree-of-freedom manipulator with rotational joints at the hip-pitch, hip-roll, and knee (see Fig.~\ref{fig:simulation:icub}).
The ankle joint is kept fixed by a position controller.
The joint limits of iCub are given in Tab.~\ref{tab:simulation:limits}.

\begin{table}[!b]
	\caption{The position and velocity limits of the iCub}
	\label{tab:simulation:limits}
	\begin{center}
		\begin{tabular}{|c||c|c|c|}
			\hline
			joint                      & hip-pitch  & hip-roll & knee\\
			\hline
			minimum position [deg]     & -45        & -20      & -120\\
			\hline
			maximum position [deg]     & 120        & 90      & 0\\
			\hline
			minimum velocity [deg/sec] & -45        & -90     & -90\\
			\hline
			maximum velocity [deg/sec] & 45         & 90      & 90\\
			\hline
		\end{tabular}
	\end{center}
\end{table}

\subsubsection{Constant desired trajectory}

\begin{figure}
	\centering
	\includegraphics[scale=0.18]{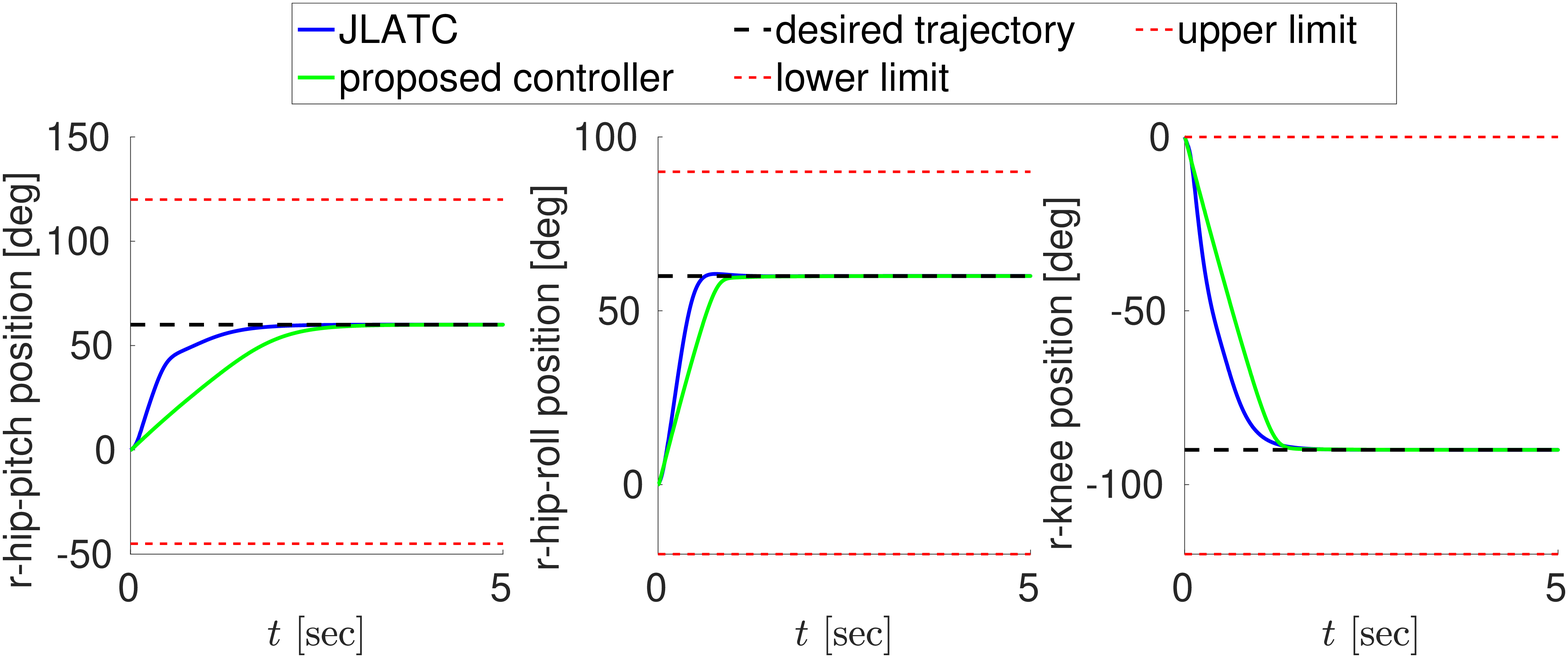}
	\includegraphics[scale=0.18]{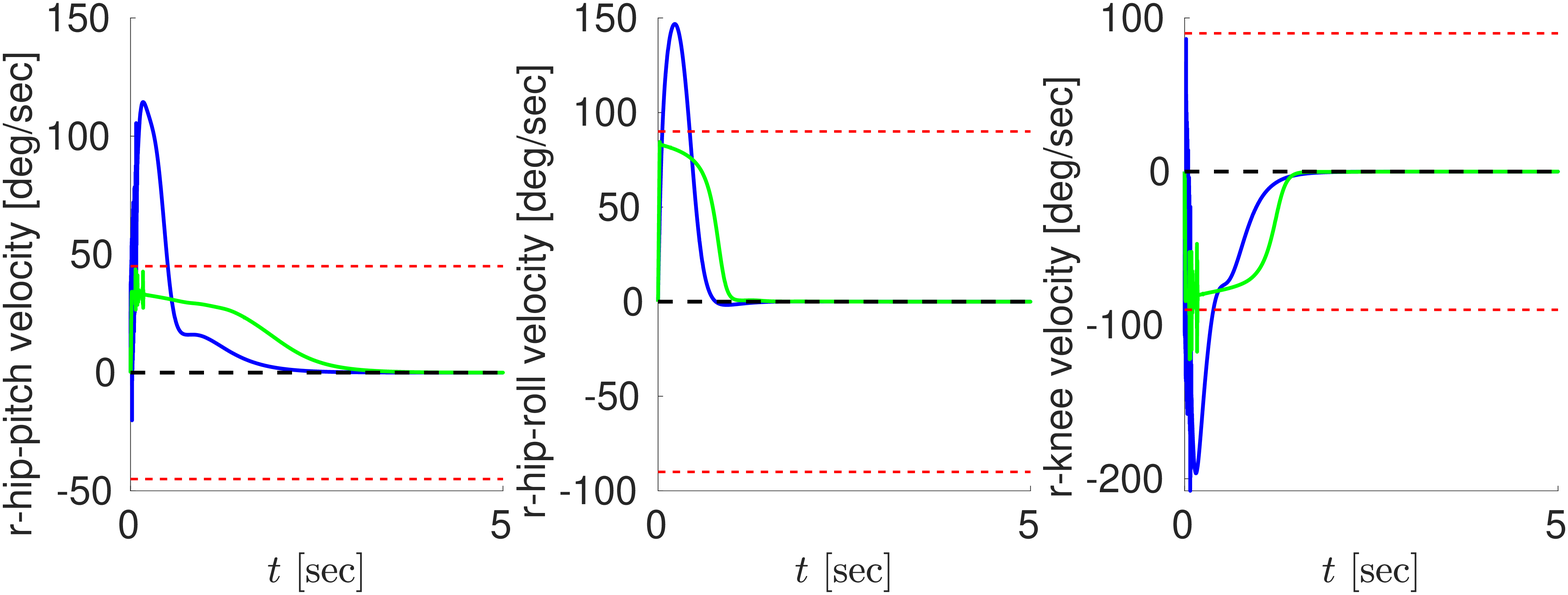}
	\includegraphics[scale=0.18]{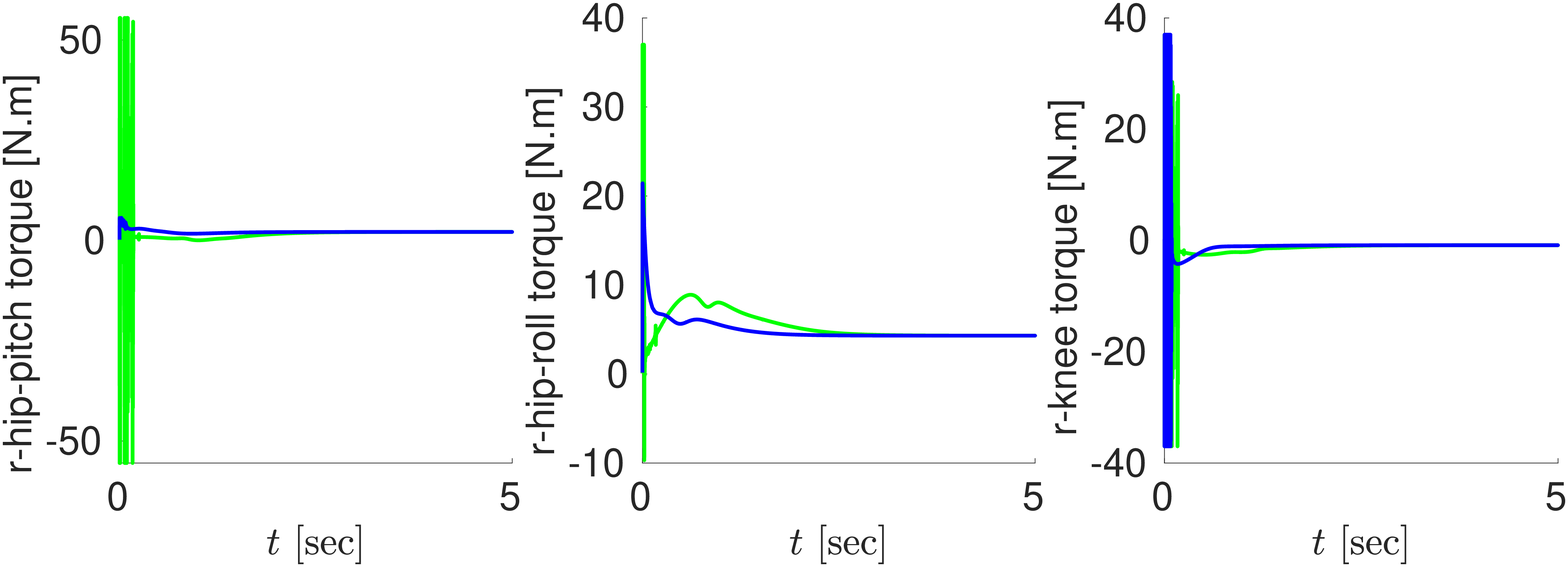}
	\caption{The joints position, velocity and torque evolution of iCub in the constant desired trajectory scenario.
		The coefficients for the proposed controller are $k_1 = 2000 I_3, k_2 = 310 I_3$, and $k_3 = 50 I_3$ and for the JLATC are $k_p = \text{diag}(8,10,6)$ and $\text{diag}(2,2,1)$.}
	\label{fig:simulation:constant}
\end{figure}

In this scenario, the robot reaches the desired joint position $\boldsymbol{q}_d = (60,60,-90)[deg]$ for the hip pitch, hip roll and knee joints respectively, from a given initial position $\boldsymbol{q} (0) = (0,0,0)[deg]$.
As can be seen in Fig.~\ref{fig:simulation:constant}, using the proposed controller, the joints trajectory converges to the desired constant value and respects the predefined position limits but, in contrast to the simulation results presented in section \ref{sec:simulation:two-link-manipulator} for the two-link manipulator, in the beginning of the simulation, joints velocity fails to respect its corresponding limits while joints torque noticeably oscillate.
Comparing to the JLATC, the proposed control architecture has the advantage that the joints velocity trajectory goes beyond its limits just in some time instances and then goes back to its feasible range.
It is worth note that the proposed controller satisfies both the joints position and velocity limits along with smooth joints torque by reducing the simulation time step.
For the sake of limited space, the simulation results for the small time step are not presented here, but are available in GitHub\footnote{https://github.com/ami-iit/paper\_pasandi\_2023\_icra-joint-limit-avoidance}.
In fact, the proposed controller is sensitive to discretized implementation.

\subsubsection{Sinusoidal desired trajectory}

\begin{figure}
	\centering
	\includegraphics[scale=0.18]{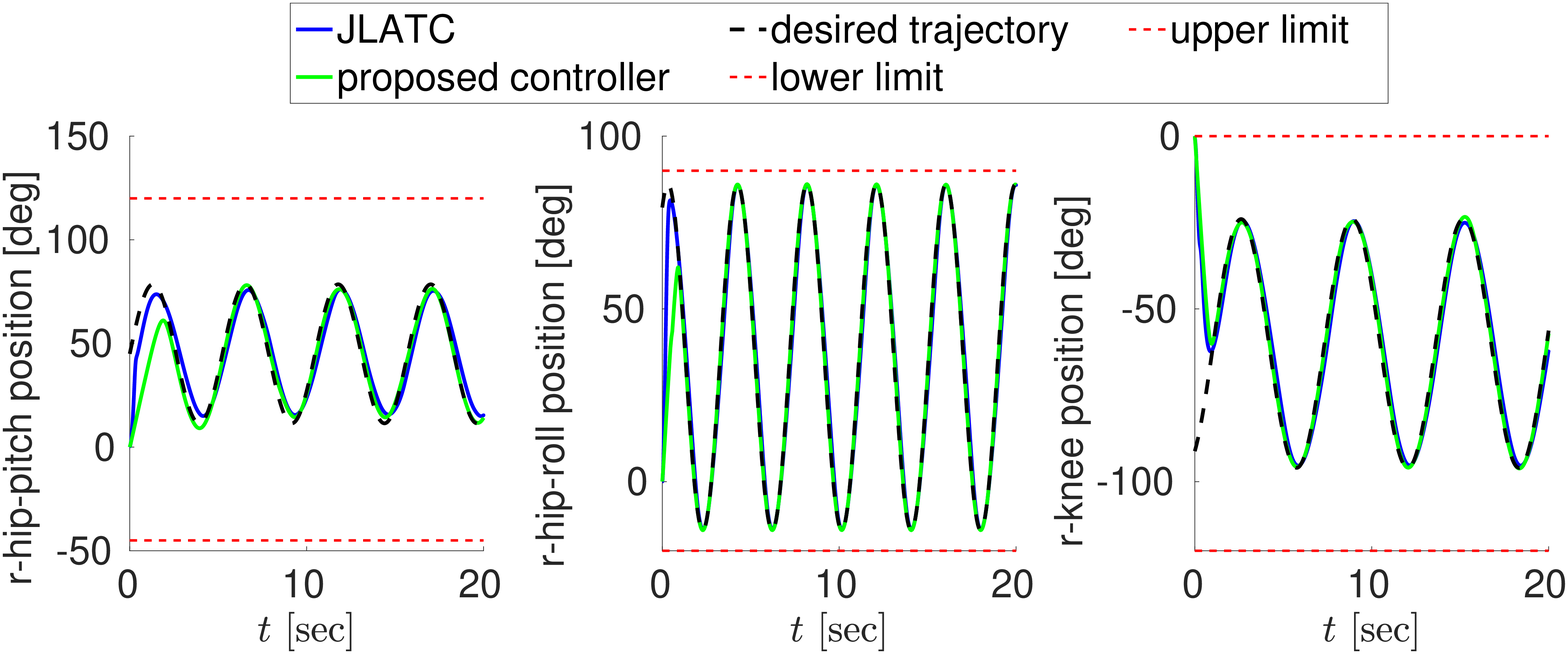}
	\includegraphics[scale=0.18]{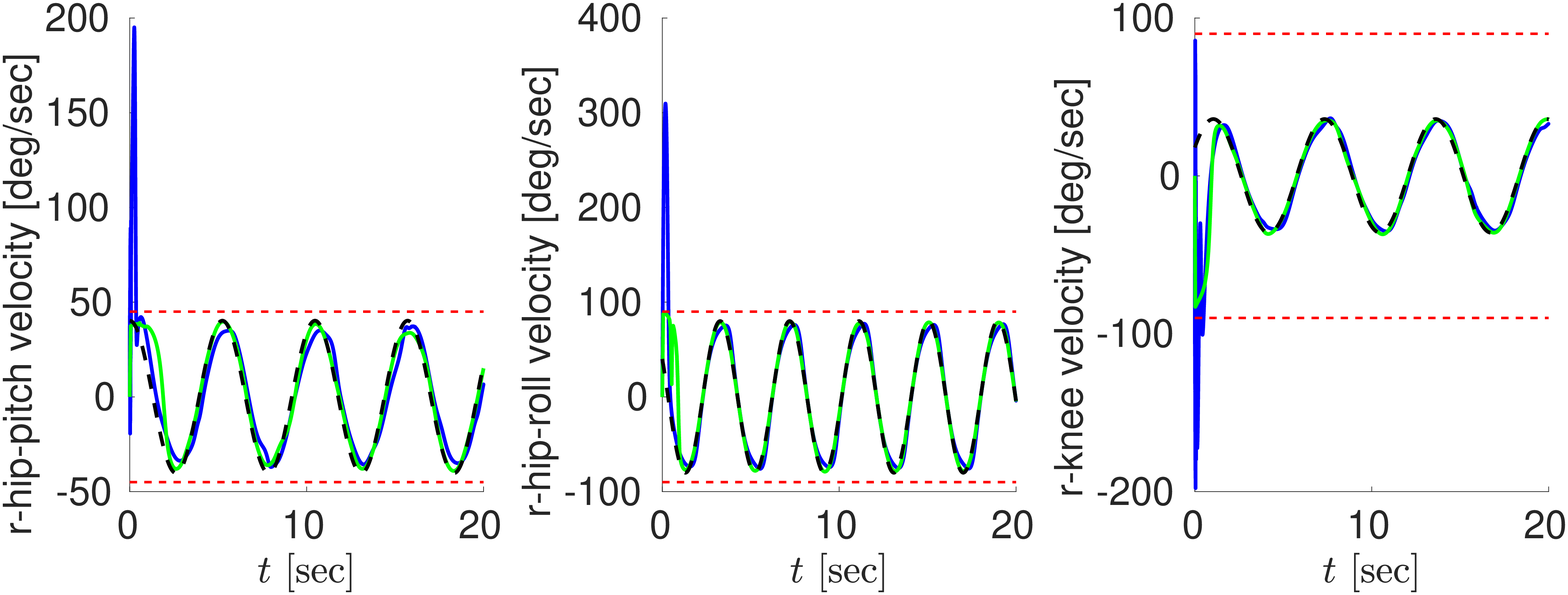}
	\includegraphics[scale=0.18]{fig/constant_joint_torque.eps}
	\caption{The joints position, velocity and torque evolution of iCub in tracking a sinusoidal desired trajectory.
		The coefficients for the proposed controller are $k_1 = 2000 I_3, k_2 = 310 I_3$, and $k_3 = 50 I_3$ and for the JLATC are $k_p = \text{diag}(8,10,6)$ and $\text{diag}(2,2,1)$.}
	\label{fig:simulation:sinusoidal}
\end{figure}

In this scenario, the robot tracks the desired sinusoidal joint position $\boldsymbol{q} = A \sin \left(\omega t + \varphi_0 \right) + b$ where $A = (33.5,50,-36) [deg]$, $\omega = (1.2,1.6,1) [rad/sec]$, $\varphi_0 = (0,\pi/3,2\pi/3) [rad]$, and $b = (45,36,-60) [deg]$.
The initial robot joint position is $(0,0,0) [deg]$.
The results are illustrated in Fig.~\ref{fig:simulation:sinusoidal}.
As can be seen, using the proposed control architecture, the joint trajectory tracks the desired sinusoidal trajectory while respecting both the joints position and velocity limits.
Instead, using the JLATC controller, the joints trajectory goes beyond the joints velocity limits.
Note that both the controllers ask high joints torques when the joints trajectory is near its position/velocity limits.

\subsubsection{Constant desired trajectory and disturbance}

\begin{figure}
	\centering
	\includegraphics[scale=0.18]{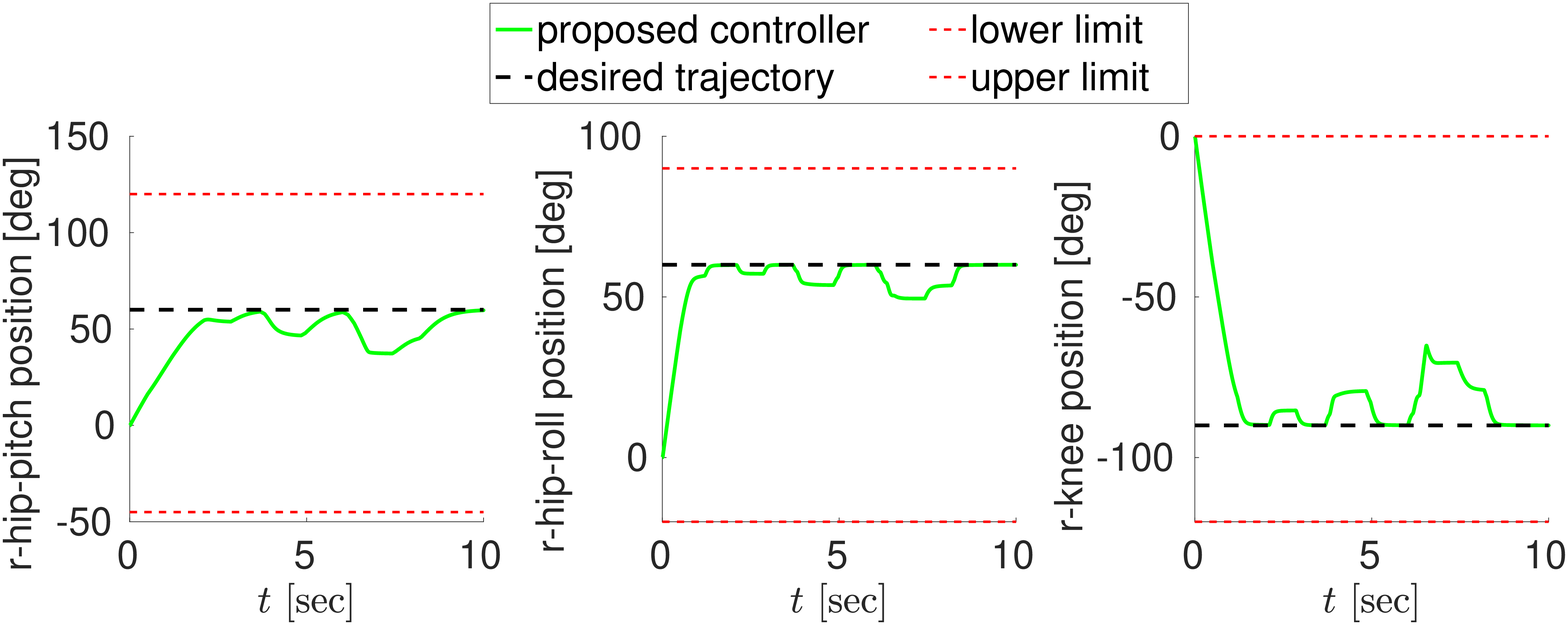}
	\includegraphics[scale=0.18]{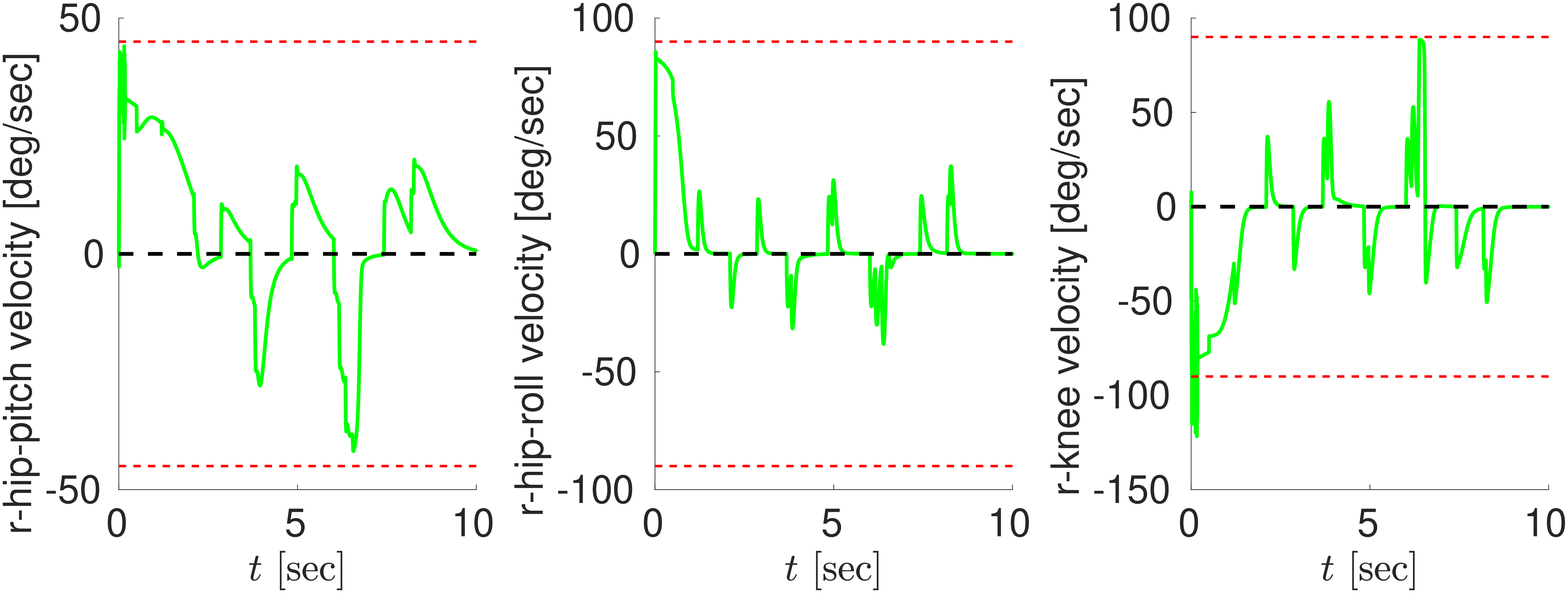}
	\includegraphics[scale=0.18]{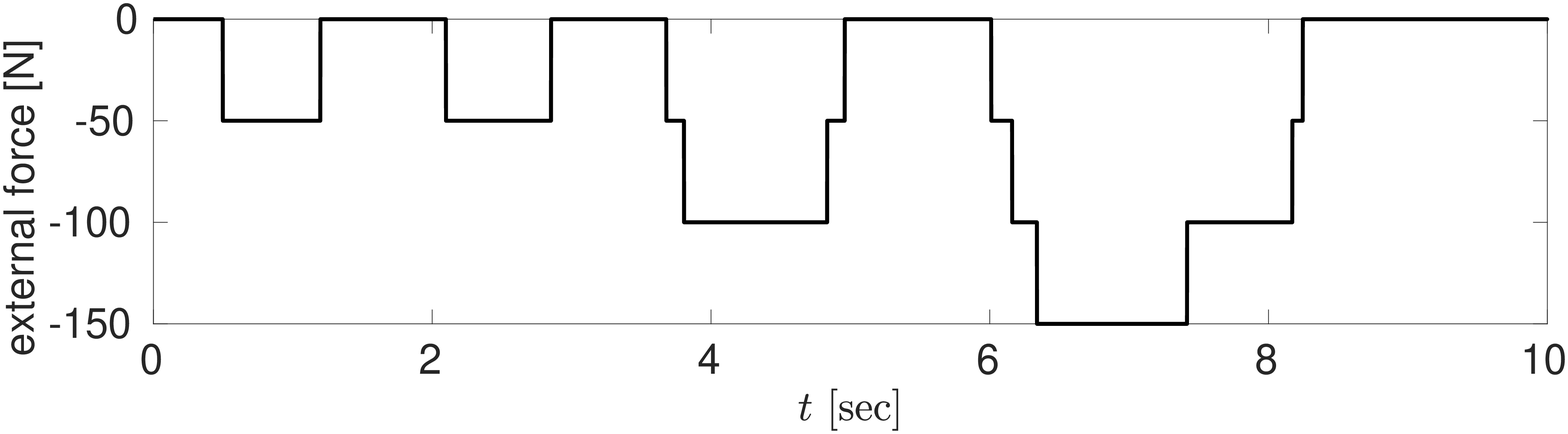}
	\caption{The joints position, velocity and torque evolution of iCub in presence of an external force.
		The coefficients for the proposed controller are $k_1 = 2000 I_3, k_2 = 310 I_3$, and $k_3 = 50 I_3$ and for the JLATC are $k_p = \text{diag}(8,10,6)$ and $\text{diag}(2,2,1)$.}
	\label{fig:simulation:constant_disturb}
\end{figure}

In this scenario, the robot reaches the constant desired joint position $(60,60,90) [deg]$ for the hip pitch, hip roll, and knee joints respectively, from the initial joint position $(0,0,0)[deg]$.
An external force in the vertical direction is applied to the sole of the robot foot.
As can be seen in Fig.~\ref{fig:simulation:constant_disturb}, the joints trajectory preserves the predefined joints position limits.
However, the joint trajectory, as explained before, goes beyond the predefined velocity limits in some time instances at the beginning of the simulation.
Later on, the joint trajectory preserves the velocity limits even in the presence of the external force.

\section{CONCLUSIONS}\label{sec:conclusion}

This paper presents a torque control architecture with joints position and velocity limits avoidance for fully actuated manipulators.
The presented architecture provides the convergence of the joints position of the robot to a desired trajectory while ensuring that the time evolution of the joints position and velocity remains within the feasible space.
The Lyapunov analysis proves the stability and convergence of the tracking error and the limitations of the joints position and velocity.
We validated the soundness of the proposed control architecture by some simulations on a simple two degrees of freedom manipulator.
We also investigated the limitations of the proposed architecture by some simulations on the humanoid robot iCub.
We observed that, in discrete implementation, the proposed controller could generate noticeable oscillatory joints torque and fail in joints velocity limit avoidance considering the discrete step time.
As future work, it is planned to reform the proposed control architecture for improving its performance in discrete implementation.

\section*{APPENDIX}

\subsection{Proof of Theorem 1}

Considering the control policy presented in \eqref{eq:control_policy}, the time derivative of $(e_{\zeta},e_{\psi})$ is
\begin{equation}\label{eq:error_dynamics}
\begin{cases}
J_{\zeta} \dot{e}_{\zeta} = \breve{\boldsymbol{\delta}}_{\dot{q}} \left( \tanh ( \boldsymbol{\psi} ) - \tanh \left( \boldsymbol{\psi}_r \right) \right), \\
\dot{e}_{\psi} + k_2^{-1} k_1 \dot{e}_{\zeta} = - k_2 e_{\zeta} - k_3 e_{\psi}. 
\end{cases}
\end{equation}
To prove the asymptotic stability of $(e_{\zeta},e_{\psi}) = (0,0)$, we consider the following candidate Lyapunov function
\begin{equation*}
V = \dfrac{1}{2}
\begin{pmatrix} e_{\zeta}^{\top} & e_{\psi}^{\top} \end{pmatrix}
P
\begin{pmatrix} e_{\zeta} \\ e_{\psi} \end{pmatrix}
= \dfrac{1}{2} 
\begin{pmatrix} e_{\zeta}^{\top} & e_{\psi}^{\top} \end{pmatrix}
\begin{pmatrix} k_1 & k_2 \\ k_2 & k_3 \end{pmatrix} 
\begin{pmatrix} e_{\zeta} \\ e_{\psi} \end{pmatrix}.
\end{equation*}
According to Theorem 1, $k_1$ and $k_3-k_2 k_1^{-1} k_2$ are diagonal positive definite matrices, and thus $P$ is a symmetric positive definite matrix based on Schur complement theorem \cite{zhang2006schur}.
As a result, $V$ is a positive definite function where $V = 0$ iff $ \left( e_{\zeta} , e_{\psi} \right) = (0,0)$.

The time derivative of $V$ along \eqref{eq:error_dynamics} is
\begin{equation*}
\begin{aligned}
\dot{V} = &\left( e_{\zeta}^{\top} k_1 + e_{\psi}^{\top} k_2 \right) \dot{e}_{\zeta} + \left( e_{\zeta}^{\top} k_2 + e_{\psi}^{\top} k_3 \right) \\
& \left( - k_2^{-1} k_1 \dot{e}_{\zeta} - k_2 e_{\zeta} - k_3 e_{\psi} \right) \\
= & -\left( \psi - \psi_r \right)^\top \left( k_3 -k_2 k_1^{-1} k_2 \right) k_2^{-1} k_1 J_{\zeta}^{-1} \boldsymbol{\delta}_{\dot{q}} \\
& \left( \tanh (\psi) - \tanh \left(\psi_r \right) \right) - \left| k_2 e_{\zeta} + k_3 e_{\psi} \right|^2.
\end{aligned}
\end{equation*}
Thanks to the diagonal property of $\left( k_3 -k_2 k_1^{-1} k_2 \right) k_2^{-1} k_1 J_{\zeta}^{-1} \boldsymbol{\delta}_{\dot{q}}$, we observe that $\dot{V}$ is negative definite.
Hence, $(e_{\zeta},e_{\psi}) = (0,0)$ is asymptotically stable based on the Lyapunov stability theorem \cite{khalil2002nonlinear}.
In conclusion, 
\begin{itemize}
	\item $(e_{\zeta},e_{\psi}) $ is bounded, and thus $(\boldsymbol{\zeta},\boldsymbol{\psi})$ is bounded if $(\boldsymbol{q}_d,\dot{\boldsymbol{q}}) \in \left(\mathcal{Q}_q,\mathcal{Q}_{\dot{q}}\right)$.
	\item $(e_{\zeta},e_{\psi})$ converges to $(0,0)$, and thus $(\boldsymbol{\zeta},\boldsymbol{\psi})$ converges to $(\boldsymbol{\zeta}_d,\boldsymbol{\psi}_d)$.
\end{itemize}
Thus, $(\boldsymbol{\zeta}_d,\boldsymbol{\psi}_d)$ is an asymptotically stable trajectory of the closed loop system in terms of Lyapunov stability \cite{khalil2002nonlinear}.
Finally, since $V$ is radially unbounded, $(\boldsymbol{\zeta}_d,\boldsymbol{\psi}_d)$ is globally asymptotically stable.

\bibliographystyle{IEEEtran}
\bibliography{ref}

\end{document}